\documentclass[10pt,letterpaper]{article}

\usepackage{cogsci}
\usepackage{pslatex}
\usepackage{apacite}
\usepackage{graphicx}

\title{Computational evolution of decision-making strategies}
 
\author{{\large \bf Peter Kvam (kvampete@msu.edu)} \\
  Center for Adaptive Rationality, Max Planck Institute for Human Development \\
  Lentzeallee 94, 14195 Berlin, Germany \\
  \AND{\large \bf Joseph Cesario (cesario@msu.edu)} \\
  Department of Psychology, Michigan State University\\
  316 Physics Rd, East Lansing, MI 48824, USA \\
  \AND {\large \bf Jory Schossau (jory@msu.edu)} \\
  Department of Computer Science and Engineering, Michigan State University \\
  428 South Shaw Rd, East Lansing, MI 48824, USA
  \AND {\large \bf Heather Eisthen (eisthen@msu.edu), Arend Hintze (hintze@msu.edu)}\\
  Department of Integrative Biology, BEACON Center for the Study of Evolution in Action, Michigan State University \\
  288 Farm Ln, East Lansing, MI 48824, USA\\
  }

\begin{document}

\maketitle

\begin{abstract}
Most research on adaptive decision-making takes a strategy-first approach, proposing a method of solving a problem and then examining whether it can be implemented in the brain and in what environments it succeeds. We present a method for studying strategy development based on computational evolution that takes the opposite approach, allowing strategies to develop in response to the decision-making environment via Darwinian evolution. We apply this approach to a dynamic decision-making problem where artificial agents make decisions about the source of incoming information. In doing so, we show that the complexity of the brains and strategies of evolved agents are a function of the environment in which they develop. More difficult environments lead to larger brains and more information use, resulting in strategies resembling a sequential sampling approach. Less difficult environments drive evolution toward smaller brains and less information use, resulting in simpler heuristic-like strategies.

\end{abstract}

\section{Introduction}

Theories of decision-making often posit that humans and other animals follow decision-making procedures that achieve maximum accuracy given a particular set of constraints. Some theories claim that decision-making is optimal relative to the information given, involving a process of maximizing expected utility or performing Bayesian inference \cite{bogacz_etal_2006,griffiths_tenenbaum_2006,vonneumann_morgenstern_1944}. Others assume that behavior makes trade-offs based on the environment, tailoring information processing to achieve sufficient performance by restricting priors \cite{briscoe2011conceptual},  ignoring information  \cite{gigerenzer_todd_1999}, or sampling just enough to satisfy a particular criterion \cite{link_heath_1975,ratcliff_1978}. In most cases, mechanisms underlying the initial development of these strategies are assumed -- either explicitly or implicitly -- to be the result of natural and artificial selection pressures.

In cognitive science research, however, the evolution of a strategy often takes a back seat to its performance and coherence. The clarity and intuitiveness of a theory undoubtedly play an immense role, as does its ability to explain and predict behavior, but whether or not a strategy is a plausible result of selection pressures is rarely considered. To be fair, this is largely because the process of evolution is slow, messy, and often impossible to observe in organisms in the lab. Fortunately, recent innovations in computing have enabled us to model this process with artificial agents.  In this paper, we propose a method of studying the evolution of dynamic binary decision-making using artificial Markov brains \cite{edlund2011integrated,marstaller2013evolution,olson2013predator} and investigate the evolutionary trajectories and ultimate behavior of these brains resulting from different environmental conditions.

In order to demonstrate the method and investigate an interesting problem, we focus on the simple choice situation where a decision-maker has to choose whether the source of a stimulus is 'signal' $S$ or 'noise' $N$ (for preferential decisions, nonspecific choices $A$ or $B$ can be substituted). A similar decision structure underlies a vast array of choices that people and other animals make, including edible/inedible, healthy/sick, safe/dangerous, and so on.  The task requires a decision-maker to take in and process information over time and make a decision about which source yielded that information. However, the decision maker is free to vary the amount of information it uses and processing it applies, and different theories make diverging predictions about how each of these should vary.  On one hand, it may be more advantageous to use every piece of information received, feeding it through a complex processing system in order to obtain maximum accuracy.  On the other, a simpler processing architecture that ignores information may be sufficient in terms of accuracy and more robust to random mutations, errors, or over-fitting.

\subsection{More complex models}

Many of the most prominent complex decision-making models fall under the sequential sampling framework \cite{bogacz_etal_2006,link_heath_1975,ratcliff_1978}.  These models assume that a decision-making agent takes or receives samples one by one from a distribution of evidence, with each sample pointing toward the signal or noise distribution. They posit that agent combines samples to process information, for example by adding up the number favoring $S$ and subtracting the number favoring $N$.  When the magnitude of this difference exceeds a criterion value $\theta$ (e.g. larger than 4 / smaller than -4), a decision is triggered in favor of the corresponding choice option ($+\theta \Rightarrow S, -\theta \Rightarrow N$).  This strategy implements a particular form of Bayesian inference, allowing a decision-maker to achieve a desired accuracy by guaranteeing that the log odds of one hypothesis ($S$ or $N$) over the other is at least equal to the criterion value. 

In these models, each piece of information collected is used to make a decision. Although organisms may not literally add and subtract pieces of information, we should expect to observe two characteristics in organisms that implement these or similar strategies. First, they should be relatively complex, storing the cumulative history of information to make their decisions. Second, they should give each piece of information they receive relatively equal weight, spreading out the weights assigned to information across a long series of inputs.

\subsection{Less complex models}

Toward the other end of the spectrum of model complexity are heuristics which deliberately ignore information in order to obtain better performance in particular environments \cite{brandstatter_etal_2006,gigerenzer_brighton_2009,gigerenzer_todd_1999}. Many of these strategies are non-compensatory, meaning that they terminate the use of information as soon as one piece of evidence clearly favors either $S$ or $N$. Accordingly, a decision maker can have a relatively simple information processing architecture, as it can just copy incoming information to its output indicators to give an answer.  Some of these require ordinal information about different sources of information and their validity, resulting in increased complexity \cite{dougherty_etal_2008}, but for the current problem we assume that all information comes from a single source.  

As a result of the relatively simple architecture and one-piece decision rules, we can expect to observe two characteristics in organisms that implement strategies similar to these heuristics. First, they should have relatively simple information processing architectures, favoring short and robust pathways that do little integration. Second, they should appear to give the most weight to the last piece(s) of information they receive before making their decision, yielding a relationship between the final decision and the sequence of inputs that is heavily skewed to the most recently received inputs.

Of course, the real behavior of artificially evolved organisms will probably lie somewhere along the spectrum between these two poles. However, we can compare the relative leanings of different populations of organisms by varying the characteristics of the environments in which they evolve.  We next describe the decision-making task and manipulations in more detail.

\section{Methods}

We were interested in examining the strategies and evolutionary trajectories that digital agents took to solve a simple dynamic decision-making problem.  To do so, we developed a binary decision-making task for the agents to solve. The fitness of an agent was defined as the number of correct decisions it made over 100 trials of the task, and the probability that it would reproduce was determined by this fitness value. Note that fitness was determined by the number of correct answers, reflecting agents' propensity to respond together with their accuracy when they did respond - there was no fitness penalty or cost for agent complexity. Formally, the probability that it generated each child of the next generation was given by its fitness divided by the total fitness across the total population (roulette wheel selection). An agent reproduced to the next generation by creating a copy of itself with random mutations. Over the course of 10,000 generations, this selection and mutation process led to evolution of agents that could successfully perform the task, and enabled us to analyze the strategies that the evolved agents ultimately developed.

\subsection{Decision task}

The task that the agents had to solve was a binary decision problem, where they received information from one source $S$ or another $N$.  The information coming from either source included two binary numbers, and therefore could yield any of the inputs [00], [01], [10], or [11].  Source $S$ would yield primarily 0s on the left and 1s on the right, and source $N$ would yield primarily 1s on the left and 0s on the right.  The exact proportion of these inputs was varied in order to alter the difficulty of the task.  For example, an easy $S$ stimulus would give 90\% 0s (10\% 1s) on the left, and 90\% 1s (10\% 0s) on the right.  The two inputs were independent, so this would ultimately give 81\% [01] inputs, 9\% [11], 9\% [00], and 1\% [10].  In a more difficult environment, an $S$ stimulus might have 60\% 0s on the left and 60\% 1s on the right, yielding 36\% [01], 24\% [11], 24\% [00], and 16\% [10].  For an $N$ stimulus, the possible inputs would be the same, but the frequency of [01] and [10] inputs would be flipped (i.e. more 1s on the left and 0s on the right).  These frequencies were not shown to the agents at the start of each trial. Instead, each trial started with a random frequency of 50\%, increasing each consecutive step by 1\% until the target frequency was reached.  This was done in part to emulate how agents encounter stimuli in real situations (i.e. stimuli progressively come into sensory range, increasing in strength over time rather than simply appearing), but also to avoid 'sticking' at a local maximum where agents simply copy their first input to outputs.

The target frequency of 1s and 0s was manipulated to be $60-90\%$ (in 5\% increments), resulting in 7 difficulty levels for different populations of agents.

For each decision, the agents received up to 100 inputs.  Each new input constituted one time step during which the agent could process that information.  If an agent gave an answer by signaling [01] to indicate $S$ or [10] to indicate $N$ (see below), then the decision process would come to a halt, where no new inputs would be given and the agent would be graded on its final answer. An agent received 1 point toward its fitness if it gave the correct answer or 0 points if it was incorrect or if it failed to answer before 100 inputs were given.

In addition to the difficulty manipulation, we included a ``non-decision time'' manipulation, where an agent was not permitted to answer until $t$ time steps had elapsed (i.e. the agent had received $t$ inputs).  This number $t$ was varied from 10 to 50 in 5-step increments, yielding 9 levels of non-decision time across different environments. Increasing $t$ tended to make agents evolve faster, as longer non-decision time tended to allow agents to more easily implement strategies regardless of difficulty level.

\begin{figure}[h] 
  \centering
    \includegraphics[width=0.5\textwidth]{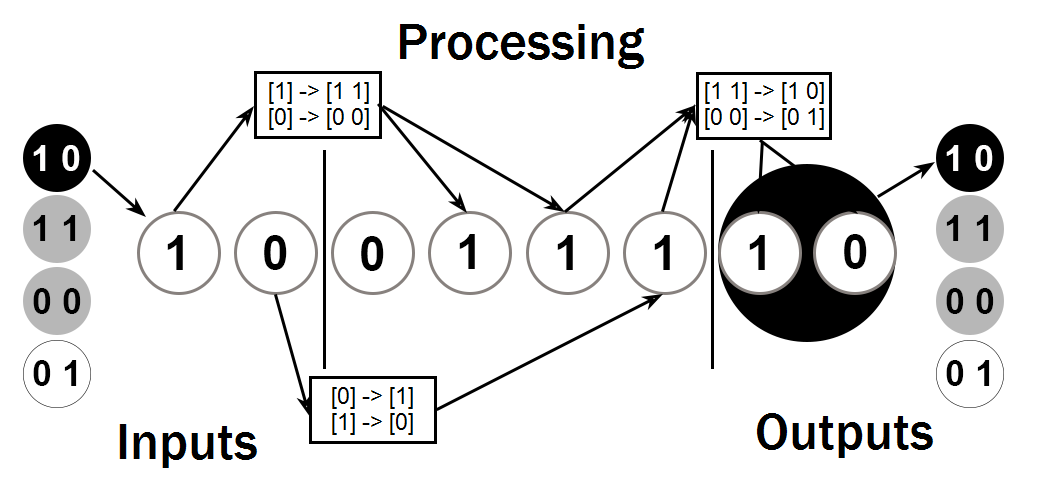}
    \caption{Diagram of the structure of a sample Markov brain with input, processing, and output nodes (circles) with connecting logic gates (rectangles).  Each gate contains a corresponding table mapping its input values (left) to output values (right).  Note that our actual agents had twice the number of nodes shown here available to them.}
    \label{Fig1}
\end{figure}

\subsection{Markov brain agents}

The Markov brain agents \cite{edlund2011integrated,marstaller2013evolution,olson2013predator} consisted of 16 binary nodes and of directed logic gates that moved and/or combined information from one set of nodes to another (see Figure \ref{Fig1}).  Two of these nodes (1 and 2) were reserved for inputs from the environment, described above.  Another two (15 and 16) were used as output nodes.  These output nodes could show any combination of two binary values.  When they did not read [01] (indicating $S$), or [10] (indicating $N$), the agents were permitted to continue updating their nodes with inputs until time step 100.  To update their nodes at each time step, the agents used logic gates (represented as squares in Figure \ref{Fig1}, which took $x$ node values and mapped them onto $y$ nodes using an $x \times y$ table.  

The input nodes, table, and output nodes for these gates were all specified by an underlying genetic code that each Markov brain possessed.  Point, insertion, or deletion mutations in the genetic code would cause them add / subtract inputs to a gate, add / subtract outputs, or change the mappings in the gate tables (e.g. it could change between any of the gates shown in Figure \ref{Fig1}).  This code consisted of 2000-200,000 'nucleotides' and included mutation rates of 0.005\% point mutations, 0.2\% duplication mutations, and 0.1\% deletion mutations, consistent with previous work \cite{edlund2011integrated,marstaller2013evolution,olson2013predator}. More precisely, logic gates are specified by 'genes' within this genetic code.  Each gene consists of a sequence of nucleotides, numbered 1-4 to reflect the four base nucleotides present in DNA, and starts with the number sequence '42' followed by '213' (start codon), beginning at an arbitrary location within the genome. Genes are typically about 300 nucleotides long and can have 'junk' sequences of non-coding nucleotides between them, resulting in the large size of the genomes. 

The first generation of Markov brain agents in each population was generated from a random seed genome. The first 100 agents were created as random variants of this seed brain using the mutation rates described above, resulting in approximately $20-30$ random connections per agent. These 100 agents each made 100 decisions, and were selected to reproduce based on their accuracy. This process was repeated for each population for 10,000 generations, yielding 100 agents per population that could perform the decision task.

\subsection{Data}

For each of the 63 conditions (7 difficulty levels $\times$ 9 non-decision times), we ran 10,000 generations of evolution for 100 different sub-populations of Markov brains, giving 6300 total populations. From each of these populations, a random organism was chosen and its line of ancestors was tracked back to the first generation. This set of agents from the last to the first generation is called the line of decent (LOD). For each of the 100 replicates per experimental conditions, all parameters (such as fitness) of agents on the LOD were averaged for each generation.

In each of these LODs, we tracked the average number of connections between nodes (see Figure \ref{Fig1}) that agents had in each condition and each generation.  We refer to this property of the agents as ``brain size'' --- the analogous properties in an organism are the number and connectivity of neurons --- and we show its evolutionary trajectory in Figure \ref{Fig2}.

Finally, we took a close look at the behavior of generation 9970 -- this is near the end to ensure that the generation we examined could solve the task, but slightly and somewhat arbitrarily removed from generation 10,000 to ensure that agents in this generation weren't approaching one of the random dips in performance (i.e. random mutations from this generation were less likely to be deleterious than more recent ones). For these agents, we examined each trial to see what information they received at each time step, which step they made their decision, and which decision they made (coded as correct or incorrect). This allowed us to examine the relationship between the inputs they received and the final answer they gave, giving an estimate of the weight they assigned to each new piece of information.

\subsection{Materials}

The agents, tasks, and evolution were implemented in C++ using Visual Studio Desktop and Xcode, and the full evolution simulations were run at Michigan State University's High Performance Computing Center.

\section{Results}

With the exception of high difficulty, low non-decision time conditions, most populations and conditions of agents were able to achieve essentially perfect accuracy on the decision task after 10,000 generations. However, the strategies implemented by each population varied heavily by condition.

It is perhaps worth noting at this point the tremendous amount of data that our approach yields. Each condition consisted of 100 populations of 100 agents that made 100 decisions each generation, yielding 10,000 agents and 1 million decisions per generation per condition. This tremendous sample size renders statistical comparisons based on standard error, for example, essentially moot. For this reason, we present mostly examples that illustrate important findings rather than exhaustive statistical comparisons.

\subsection{Brain size}

\begin{figure}[h]
  \centering
    \includegraphics[width=0.5\textwidth]{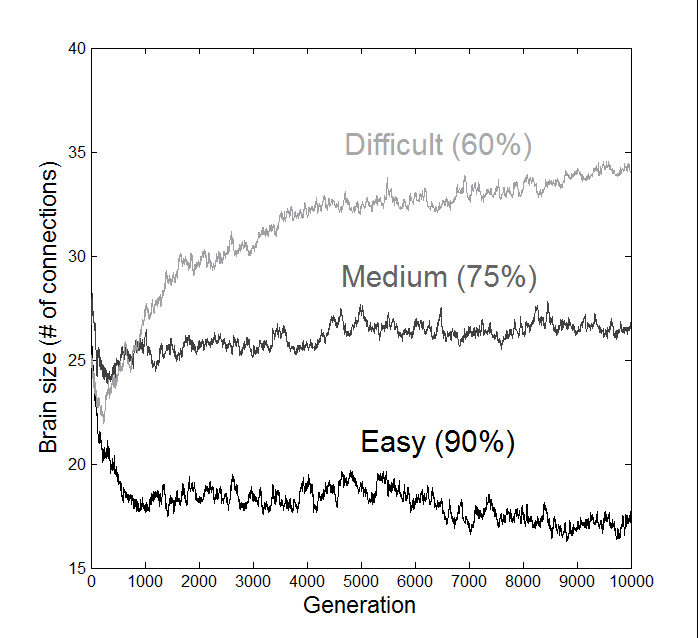}
    \caption{Mean number of connections in agent brains across generations for three levels of task difficulty. For the sake of comparison, the trajectories shown are all from populations with a non-decision time of 40 steps}
    \label{Fig2}
\end{figure}

Final brain size (number of connections among nodes) varied as a function of both stimulus difficulty and non-decision time.  We focus primarily on high non-decision time conditions, as many of the low non-decision time populations --- particularly in the difficult stimuli conditions --- were unable to achieve the high performance of other groups. As Figure \ref{Fig2} shows, agents faced with the easiest conditions (10-15\%) tended to have the smallest final brain size, with means of around $15-20$ connections. Agents faced with medium difficulty environments evolved approximately $25-30$ connections, and agent brain size in the most difficult conditions approached 35 connections and appeared to still be climbing with further generations.

Perhaps more interesting, though, is the evolutionary trajectory that each of the populations in these conditions took.  As shown, each group started with 25-30 connections in the initial generation, and in all of them the number of connections initially dropped for the first 200-400 generations.  After that, however, the conditions appear to diverge, with the agents in the easy conditions losing even more connections, agents in the medium conditions staying approximately level, and agents in the difficult conditions adding more and more connections.

\subsection{Strategy use}

\begin{figure}[h]
  \centering
    \includegraphics[width=0.5\textwidth]{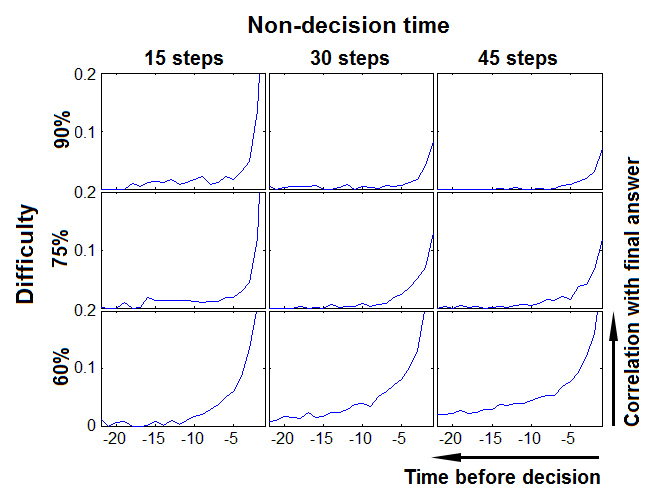}
    \caption{Example correlations between inputs and final decision for easy (blue), medium (purple) and difficult (red) conditions. The trajectories are time-locked on the final answer on the right side, so the last piece of information an agent received is the rightmost value, and to the left is moving backward through the trajectory.}
    \label{Fig3}
\end{figure}

In order to examine the pattern of information use in the agents, we additionally examined the relationship between each piece of information received and the final answer given. We did so by taking the series of inputs (e.g. [00],[11],[01],[01],[11]) and assigning each one a value - information favoring $S$ ([01] inputs) was assigned a value of +1, information favoring $N$ ([10] inputs) was assigned a value of $-1$, and others ([00] and [11]) were assigned a value of 0. Answers favoring $S$ were also given a value of +1 and answers favoring $N$ a value of $-1$.  Doing so allowed us to track the sequence of $-1$, 0, +1 --- which we refer to as the \textit{trajectory} --- leading to the decision and to correlate this with the final +1 or $-1$ answer. The result of this analysis for the example conditions is shown in Figure \ref{Fig3}.

As shown, the trajectory correlations in the more difficult conditions tend to be flatter than those in the easy conditions, and final answers tend to correlate with a longer history of inputs. This indicates that these agents were assigning more similar weight to each piece of information they use, utilizing the full history of inputs they had received rather than just the final piece.  Note that all agents appeared to use the most recent pieces of information more heavily.  This will be the case for almost any model that generates the data, as the last pieces of information tend to be those that trigger the decision rule -- for example, in sequential sampling this will be the piece of information that moves the evidence across the threshold -- and as such will always be highly correlated with the final answer. \footnote{However, since it can sometimes take several updates / time steps to move a 'trigger' input through the brain to the output nodes, the final piece of information will not always be perfectly correlated with the output.}

Information use also varied somewhat across levels of non-decision time, but its effect was not particularly pronounced except in the more difficult conditions (e.g. 60-70\%). However, this effect is largely a consequence of agent populations' failure to evolve to perform the task as well when stimulus discriminability and non-decision time were low. For example, agents in the difficult, short non-decision time condition (red in left panel of Figure \ref{Fig3}) attained accuracy of only ~82\%, compared to 95+\% in other conditions. Higher difficulty still led to larger brains and a longer history of processing in these conditions, but its effect was less pronounced.  Therefore, high values of non-decision time apparently made it easier to evolve complex strategies, likely because agents were exposed to more information before making their decisions.

\section{Discussion}

While agents' strategies spanned a range of complexity, more difficult environments pushed them toward more complex strategies resembling sequential sampling while easier environments led to strategies more similar to non-compensatory heuristics.  Therefore, both sequential sampling and heuristics seem to be strategies that could plausibly result from different environmental demands.  However, our results run counter to the idea that heuristics are invoked when decisions are particularly difficult or choice alternatives are not easily distinguished \cite{brandstatter_etal_2006}.  

The final strategies may not support the claim that organisms are primarily heuristic decision-makers \cite{gigerenzer_brighton_2009}, but it still lends credence to the premise of \textit{ecological rationality} on which many heuristics are based.  This approach suggests that different environments (choice ecologies) lead to different decision-making strategies rather than a one-size-fits-all process. It is certainly plausible that agents in environments with mixed or changing difficulty levels converge on a single strategy, but for the moment it seems that multiple strategies can be implemented across multiple choice environments.

While difficult conditions led to larger brains and more information processing, perhaps a more critical finding is that simpler choice environments led to simpler decision strategies and architectures. While this may initially seem like the other side of the same coin, this result is particularly interesting because we did not impose any penalties for larger brains. Although other researchers have suggested that metabolic costs limit the evolution of large brains \cite{isler_schaik_2006,laughlin1998metabolic} and can be substantial in real brains \cite{kuzawa2014metabolic}, they were not necessary to drive evolution toward smaller brains.

Instead, we suspect that the drop in brain size is a result of the agents' response to mutations, or the \textit{mutation load} imposed by the size of its genome. For example, a random mutation in the genome that connects, disconnects, or re-maps a gate is more likely to affect downstream choice-critical elements of a brain that uses more nodes and connections to process information (has a higher mutation load), particularly if it has a larger ratio of coding to non-coding nucleotides.  In this case, a smaller brain would be a tool for avoiding deleterious mutations to the information processing stream.  Alternatively, the minimum number of nodes and connections required to perform the task is likely lower in the easier conditions than in the more difficult ones, so mutations that reduce brain size and function might be able to persist in the easier but not the more difficult conditions. In either case, it is clear that a larger brain does not offer sufficient benefits in the easier conditions to overcome the mutation load it imposes.

Another potential risk of having a larger brain is the chance of a random mutation preventing information from reaching the output nodes -- with a longer chain of processing nodes being easier to interrupt or confuse than a shorter one.  While the agents in more difficult conditions were evidently able to overcome such a possibility (usually answering within ~20 steps of the end of non-decision time), it may be a barrier that required substantial fitness rewards to cross, which were not present in the easier conditions.

We hesitate to make claims that are too broad given the scope of our study, but the finding that brain size can be limited by mutation load is provoking. This may explain why systems that are subject to mutations and selection pressures -- including neurons and muscle cells -- are reduced when they are unused, even when the energetic costs of maintaining the structure appear to be low. It seems a promising direction for future research to examine in-depth how mutation rate and robustness contribute to organisms' fitness above and beyond the costs associated with metabolism.

\subsection{Approach}

We hope to have presented a method for examining questions regarding adaptation and evolution that often arise in cognitive science and psychology.  Whereas previous studies have worked from a particular strategy and examined the choice environments in which it succeeds, we present a way of answering questions about how the environment can shape the evolution of a strategy. The strategies resulting from this computational evolution approach are adaptive, easily implemented in the brain, and the result of realistic natural selection pressures. Additionally, we have shown that this approach is capable of addressing important questions about existing models of simple dynamic decisions, though it could undoubtedly shed light on an array of related problems.

Of course, there are limitations to this approach, many of which are computational. The agents we used had only 16 nodes, 4 of which were reserved for inputs and outputs, meaning that only 12 could be used for storing (memory) and processing information. Although more nodes could be added -- and certainly an accurate model of even very simple nervous systems would have many times more -- this would severely slow down the steps required for evolution. It might also lead to problems that are analogous to the over-fitting that occurs when more parameters are added to a model, though this is itself a question worth exploring.

\subsection{Conclusions}

In this paper, we presented a computational evolution framework that could be used to examine how environments lead to different behaviors. This framework allowed us to examine the strategies that might have arisen in organisms to address the problem of dynamic decision-making, where agents receive information over time and must somehow use this input to make decisions that affect their fitness. 

We found that both the evolutionary trajectory and the strategies ultimately implemented by the agents are heavily influenced by the characteristics of the choice environment, with the difficulty of the task being a particularly notable influence. More difficult environments tended to encourage the evolution of complex information integration strategies, while simple environments actually caused agents to decrease in complexity, perhaps in order to maintain simpler and more robust decision architectures. They did so despite no explicit costs for complexity, indicating that mutation load may be sufficient to limit brain size.

Finally, we discussed these results in the context of existing models of human decision-making, suggesting that both non-compensatory strategies such as fast and frugal heuristics \cite{gigerenzer_todd_1999} and complex ones such as sequential sampling \cite{link_heath_1975} may provide valid descriptions -- or at least serve as useful landmarks -- of the strategies implemented by evolved agents. In doing so, we provided evidence that strategy use is environment-dependent, as different decision environments led to different patterns of information use. More generally, we have shown that a computational evolution approach integrating computer science, evolutionary biology, and psychology is able to provide insights into how, why, and when different decision-making strategies evolve.

\section{Acknowledgments}

This work was supported by Michigan State University's High Performance Computing Facility and the National Science Foundation under Cooperative Agreement No. DBI-0939454 and Grant No. DGE-1424871.

\bibliographystyle{apacite}

\setlength{\bibleftmargin}{.125in}
\setlength{\bibindent}{-\bibleftmargin}

\setlength{\bibleftmargin}{.125in}
\setlength{\bibindent}{-\bibleftmargin}

\bibliography{CogSciBib}

\end{document}